# Over-enhancement Reduction in Local Histogram Equalization using its Degrees of Freedom

*Alireza Avanaki*

**ABSTRACT**
A well-known issue of local (adaptive) histogram equalization (LHE) is over-enhancement (i.e., generation of spurious details) in homogenous areas of the image. In this paper, we show that the LHE problem has many solutions due to the ambiguity in ranking pixels with the same intensity. The LHE solution space can be searched for the images having the maximum PSNR or structural similarity (SSIM) with the input image. As compared to the results of the prior art, these solutions are more similar to the input image while offering the same local contrast.

## 1. INTRODUCTION & BACKGROUND

Automatic contrast enhancement is a common operation on visual data that is used to reveal otherwise invisible details. Histogram equalization (HE) or specification (when the input is modified to have a specified histogram, which is flat in HE) are widely used for this purpose. In global histogram equalization (GHE), a space-invariant transform is applied to the gray-level (GL) of each pixel[1]. This way, the GLs are redistributed so that the output has (almost) flat histogram [1]. In local (a.k.a. adaptive or block-overlapped) histogram equalization (LHE) [2-3], the above mentioned transform is space-variant; HE is performed in a sliding window, but the result is only kept for the central pixel of the window.

GHE cannot guarantee enough contrast enhancement in all regions of the image. LHE, on the other hand, captures the details of its input image well but it also generates noise (i.e., overly accentuated or non-existing details; a.k.a. over-enhancement) in the homogenous areas of the input.

The state of the art in mitigation of contrast over-enhancement is the method given in [3], where by adding a fraction of local mean, and modifying local histogram accumulation, effects between LHE and the unaltered input are obtained. Because of such modifications, the result of [3] does not have a locally equalized histogram. Hence they cannot provide the maximum local contrast enhancement offered by a LHE solution.

In this paper, we examine the LHE over-enhancement problem from a perspective acquired from exact histogram specification [8] (EHS). In EHS, the input image has to be modified so that the histogram of the result exactly matches the specified histogram. To that end, the pixels of input with the same GL (in one bin of the input histogram) may be assigned different GLs in the EHS result. In [8-9], such "bin-splitting" is avoided using auxiliary information, in addition to pixels' GL values, that establish a strict ordering of pixels. In [6], bin-splitting is performed so that the structural similarity (SSIM; see Appendix or [5]) between the EHS result and the input is maximized. In this work, we inspect the local counterpart of the bin-splitting problem. Depending on how the histogram bin containing the center pixel of the sliding window is split, the LHE result at each pixel has a degree of freedom (DoF). From such DoFs, one can select values to satisfy specific constraints or optimize a certain metric, globally or locally. We found no prior work pointing out the DoFs in the LHE result.

---

[1] We only consider grayscale visual data. Color images can be treated by the application of the proposed methods to their luminance channels.

The following is furnished in this paper. (i) The LHE problem has several valid solutions. The LHE solution space can be searched for (ii) the solution that maximizes PSNR of the result with the input, (iii) and the solution that maximizes SSIM of the result with the input. Such solutions reduce over-enhancement by preserving the similarity between the result and the original image in PSNR and SSIM senses respectively. Hence, the smooth areas of the input image are also kept smooth in the result, as much as possible.

The rest of the paper is organized as follows. Sections 2 to 4 describe the contributions listed in the last paragraph in the same order. In Section 5, the tools developed in Sections 3 and 4 are used to enhance the results of an existing over-enhancement remedy. Numerical results are given in Section 6. Section 7 concludes the paper. SSIM and its gradient are reviewed in Appendix.

## 2. LHE SOLUTION SPACE

To perform LHE, the pixels in a sliding window are sorted in ascending GL order. Then the rank of the center pixel is recorded as the result for that pixel. For example, consider the following 3x3 windows of an 8-bit grayscale image.

$$N_1 = \begin{pmatrix} 10 & 12 & 25 \\ 25 & 36 & 47 \\ 47 & 65 & 77 \end{pmatrix}, \quad N_2 = \begin{pmatrix} 10 & 12 & 25 \\ 25 & 25 & 47 \\ 56 & 65 & 25 \end{pmatrix}$$

The center pixel rank in $N_1$ is 5 since four of the nine pixels have GLs lower than 36, the center pixel GL. In $N_2$, the center pixel rank is not unique and can be assigned any rank in [3, 6] (brackets are used to show a closed integer range). For a $N_3 = const.$, the valid rank range is [1, 9]. This range may be translated to a DoF of [0, 255] (i.e., spanning all possible GLs), if the LHE result is expressed in the same bit-depth as that of the original image.

These DoFs in the LHE result, for example, can be used to reduce over-enhancement effect by maximizing the similarity of the LHE result and the original (input) image (Sections 3 and 4). This way, the over-enhanced areas, which are less similar to the input image, are forced to be more similar (i.e., less enhanced) to the input.

In the following, we derive DoFs for the LHE result. The local histogram of image $I$ at $(m,n)$ is given by [3]

$$h(m,n,\alpha) = B_\alpha(m,n) * W(m,n), \text{ with } B_\alpha(m,n) = \begin{cases} 1, & I(m,n) = \alpha \\ 0, & \text{Otherwise} \end{cases}, \text{ and} \quad (1)$$

$$W(m,n) = \begin{cases} 1, & |m| \leq w \text{ and } |n| \leq w \\ 0, & \text{Otherwise} \end{cases}.$$

'*' denotes 2-D discrete convolution. In other words, $h(m,n,\alpha)$ is the number of pixels with GL of $\alpha$ in a $(2w+1) \times (2w+1)$ window of $I$ centered on $(m,n)$. With the 'box car' kernel above, however, the convolution boils down to counting, which is simpler.

The cumulative local histogram is given by

$$H(m,n,\alpha) = \sum_{\beta=0}^{\alpha} h(m,n,\beta), \text{ for } \alpha = 0,...,L-1, \quad (2)$$

for a grayscale image with $L$ possible GLs. The rank of pixel $(m,n)$ within its $(2w+1) \times (2w+1)$ neighborhood can be considered anywhere in

$$(H(m,n,I(m,n)-1), \ H(m,n,I(m,n))], \tag{3}$$

in which (.,.] denotes an integer range open on left and closed on right. Note that when $I(m,n)$ is a unique GL in the window, $H(m,n,I(m,n)-1)+1 = H(m,n,I(m,n))$, and that $H(m,n,-1) = 0$. The LHE result is desired to be in $[0, L-1]$, onto which we linearly map the range of (3). Hence, the degrees of freedom (DoF) for the result image at $(m,n)$ is given by

$$[\mathbf{L}(m,n), \ \mathbf{U}(m,n)], \text{ with } \mathbf{L}(m,n) = \min\left(L-1, \left\lfloor L \frac{H(m,n,I(m,n)-1)}{H(m,n,L-1)} \right\rfloor\right), \text{ and} \tag{4}$$

$$\mathbf{U}(m,n) = \min\left(L-1, \left\lfloor L \frac{H(m,n,I(m,n))}{H(m,n,L-1)} \right\rfloor\right),$$

where $\lfloor . \rfloor$ denotes integer floor operator. Note that $\max_\alpha H(m,n,\alpha) = H(m,n,L-1)$ since $H(.,.,\alpha)$ is non-decreasing with respect to $\alpha$.

The number of valid LHE solutions for $I$ (i.e., the size of LHE solution space) is given by

$$\prod_{\forall (m,n)} (\mathbf{U}(m,n) - \mathbf{L}(m,n) + 1), \tag{5}$$

in which the multiplication is performed over all pixels of $I$.

By replacing $H(m,n,.)$ with $G^{-1}(m,n,H(m,n,.))$ [8], where $G(.,.,.)$ is the specified (a.k.a. target) local cumulative histogram, (3) to (5) can be used for local histogram specification.

## 3. MSE-OPTIMAL SOLUTION

Among LHE solutions, in this section, we find the ones that minimize or maximize mean square error (MSE) with the original (input) image. Such solutions show maximum or minimum similarity with the input in terms of PSNR. The latter is useful in computing a lower bound on PSNR of all LHE solutions.

For the input image $I$, the LHE solution with the best quality in MSE sense is given by

$$X_{\min \text{MSE}} = \arg\min_Y \sum_{\forall (m,n)} (Y(m,n) - I(m,n))^2 \tag{6}$$

s.t. $\mathbf{L}(m,n) \leq Y(m,n) \leq \mathbf{U}(m,n), \ \forall (m,n)$

where $\mathbf{L}$, and $\mathbf{U}$ are given by (4).

From the constraint in (6), we get

$$\mathbf{L}(m,n) - I(m,n) \leq Y(m,n) - I(m,n) \leq \mathbf{U}(m,n) - I(m,n). \tag{7}$$

To perform the minimization in (6), each term should be minimized. Thus,

$$X_{\min \text{MSE}}(m,n) = \begin{cases} \mathbf{L}(m,n), & I(m,n) < \mathbf{L}(m,n) \\ \mathbf{U}(m,n), & I(m,n) > \mathbf{U}(m,n) \\ I(m,n), & \text{Otherwise} \end{cases} \tag{8}$$

Following a similar procedure, the LHE solution with the worst quality in terms of MSE is given by

$$X_{\max MSE}(m,n) = \begin{cases} \mathbf{U}(m,n), & I(m,n) \leq \mathbf{L}(m,n) \\ \mathbf{L}(m,n), & I(m,n) \geq \mathbf{U}(m,n) \\ \arg\max_{y=\{\mathbf{L}(m,n),\mathbf{U}(m,n)\}} |y - I(m,n)|, & \text{Otherwise} \end{cases} \quad (9)$$

## 4. OPTIMIZING LHE FOR STRUCTURAL SIMILARITY

Among LHE solutions, in this section, we try to find the one with the maximum structural similarity index (SSIM) with the original image. SSIM, described in Appendix, is a full-reference image quality metric that is shown to correlate well with the perceptual quality [5].

For the input image $I$, the LHE solution with the best quality in terms of SSIM is given by

$$X_{\max SSIM} = \arg\max_Y SSIM(I,Y) \quad (10)$$

s.t. $\mathbf{L}(m,n) \leq Y(m,n) \leq \mathbf{U}(m,n), \quad \forall (m,n)$

where $\mathbf{L}$, and $\mathbf{U}$ are given by (4).

A closed-form solution to this problem may not exist. The following lemma is used in development of a suboptimal iterative solution.

**Lemma.** If $Y$ is a LHE solution, and

$$X = Y + \beta \nabla_Y SSIM(I,Y), \quad \text{with } \beta > 0 \quad (11)$$

$$Y'(m,n) = \begin{cases} \mathbf{L}(m,n), & X(m,n) < \mathbf{L}(m,n) \\ \mathbf{U}(m,n), & X(m,n) > \mathbf{U}(m,n) \\ [X(m,n)], & \text{Otherwise} \end{cases} \quad (12)$$

then $Y'$ is also a LHE solution, and the first order approximation of $SSIM(I,Y')$ is not less than $SSIM(I,Y)$. In (11), $\nabla_Y$ denotes gradient with respect to image $Y$ (i.e., $\nabla_Y = \left(\frac{\partial}{\partial y_i}\right)_{i=1,\ldots,N}$, where $y_i$ is the GL of the $i^{th}$ pixel of $Y$, and $N$ is the number of pixels in $Y$), [.] denotes rounding to the nearest integer.

**Proof.** $Y'$ is a LHE solution because (12) ensures that all pixels of $Y'$ are within LHE DoFs given bounded by $\mathbf{L}$, and $\mathbf{U}$. We show that the first-order change in SSIM caused by (11) and (12) is non-negative. Consider a pixel of $Y$, with the GL of $y$ and the updated GL of $x$, given by (11), that is assigned a GL of $y'$, given by (12). In this notation, the SSIM gradient for this pixel is given by $\beta^{-1}(x-y)$, and the change in SSIM (of $Y'$ as compared to that of $Y$) contributed by this pixel is $\Delta SSIM = \beta^{-1}(x-y)(y'-y)$.

We distinguish the following three cases: (i) $x < \mathbf{L}(m,n)$, (ii) $x > \mathbf{U}(m,n)$, (iii) $\mathbf{L}(m,n) \leq x \leq \mathbf{U}(m,n)$.

In case (i), since $Y$ is assumed to be a LHE solution, we have $\mathbf{L}(m,n) \leq y \leq \mathbf{U}(m,n)$. Hence, $x \leq y$, and $y' = \mathbf{L}(m,n) \leq y$, because of (12). Since the last two multiplicands of $\Delta SSIM$ are non-positive and $\beta > 0$, we have $\Delta SSIM \geq 0$.

In case (ii), similar to case (i), it can be shown that the last two multiplicands of $\Delta SSIM$ are non-negative; thus $\Delta SSIM \geq 0$.

In case (iii), if $y' = x = y$, then $\Delta SSIM = 0$ (non-negative). If $x > y$, then $y' = [x]$ cannot be smaller than $y$ (which is an integer), thus $y' > y$; which means that the last two multiplicands of

$\Delta$SSIM are positive, and $\Delta$SSIM $> 0$. If $x < y$, the last two multiplicands of $\Delta$SSIM become negative hence $\Delta$SSIM $> 0$. **End of proof.**

This lemma guarantees that the first order approximation of LHE solution quality (in terms of SSIM) increases (or at least doesn't decrease) for small (not too small, see Section 4.1) positive $\beta$ in Algorithm 1. In other words, Algorithm 1 is a SSIM gradient ascent in the subspace of LHE solutions. The first line in the main loop enhances the solution quality (i.e., its similarity with the input image in terms of SSIM). The inner loop projects the enhanced solution to its closest valid LHE solution, with a better (or the same) quality than the solution from the last iteration. The stopping criterion may be either (or a combination) of the following: (i) The solution quality is above a given threshold; (ii) The growth in solution quality is under a certain threshold. (iii) The number of iterations reaches a limit. The latter is used in the experiments of %Section 7.

**Algorithm 1.** SSIM-enhanced LHE solution

Input $\mathbf{L}, \mathbf{U}$ (LHE DoF bounds for $I$), and $I$.

$Y = Y_{init}$

While 1,

  $X = Y + \beta \nabla_Y \text{SSIM}(I, Y)$

  For all $(m, n)$, // Projection

  $$Y(m,n) = \begin{cases} \mathbf{L}(m,n), & X(m,n) < \mathbf{L}(m,n) \\ \mathbf{U}(m,n), & X(m,n) > \mathbf{U}(m,n) \\ [X(m,n)], & \text{Otherwise} \end{cases}$$

  End for

  If stopping crietrion is met, break

End while

Output $Y$

## 4.1 Step size selection

By using a step size that is too large, we may skip the maximum during gradient ascent. On the other hand, a step size that is too small requires more iterations for a certain increase in SSIM. In the following, we derive the step size that yields the maximum SSIM growth in each iteration. Although such a greedy approach may not necessarily lead to *the* highest quality LHE solution, it consistently enhanced the quality of the result in all of our experiments (reported in %Section 7).

The value of step size that maximizes SSIM growth in each iteration is given by:

$$\beta_{opt} = \arg\max_{\beta} \text{SSIM}(I, \text{P}(\mathbf{L}, Y + \beta \nabla_Y \text{SSIM}(I, Y), \mathbf{U})) \quad (13)$$

in which P(.,.,.) denotes the projection step (i.e., the inner loop) in Algorithm 1. This maximization problem is difficult because the projection function is not differentiable. We observed that argument of max(.) in (13) is a smooth function of $\beta$ in proximity of $\beta_0$, a rough estimate of $\beta_{opt}$ calculated below. Therefore, (13) can be solved numerically by a scalar search (e.g., Nelder-Mead method) with a starting point of $\beta_0$.

To estimate $\beta_{opt}$, we assume that $P(\mathbf{L}, Y+\beta\nabla_Y\text{SSIM}(I,Y), \mathbf{U}) \cong [Y+\beta\nabla_Y\text{SSIM}(I,Y)]$ (i.e., saturation in projection rarely occurs). Then, using a popular model of round-off, we have

$$P(\mathbf{L}, Y+\beta\nabla_Y\text{SSIM}(I,Y), \mathbf{U}) \cong Y+\beta\nabla_Y\text{SSIM}(I,Y)+E, \tag{14}$$

in which $E$ is a zero-mean uniform noise that is uncorrelated with the signal. The change in SSIM is therefore $(\beta\nabla_Y\text{SSIM}(I,Y)+E)\cdot\nabla_Y\text{SSIM}(I,Y) = \beta\|\nabla_Y\text{SSIM}(I,Y)\|^2$, since $E\cdot\nabla_Y\text{SSIM}(I,Y)$ vanishes ('.' denotes dot product). Ideally, if $Y$ becomes $I$ due to this change, we have $\beta\|\nabla_Y\text{SSIM}(I,Y)\|^2 = 1-\text{SSIM}(I,Y)$. Hence an estimate of $\beta_{opt}$ is given by

$$\beta_0 = \frac{1-\text{SSIM}(I,Y)}{\|\nabla_Y\text{SSIM}(I,Y)\|^2}. \tag{15}$$

This estimate or a fraction (typically 0.2 to 1; justified by assuming that several iterations are required for $Y$ to approach $I$) of it, can be used as a starting point of scalar search for $\beta_{opt}$. In the case that processing speed has priority over result optimality, $\beta_0$ can replace $\beta$ in Algorithm 1.

**4.2 Initialization**
Since SSIM($I,Y$) is not a convex function of $Y$, the final value of SSIM($I,Y$) depends on the starting point of gradient ascent, $Y_{init}$. We observed that two good choices of $Y_{init}$ are $\left[\dfrac{\mathbf{L}+\mathbf{U}}{2}\right]$, and $X_{min\,MSE}$, given by (8). With the greedy growth step size selection (Section 4.1), both choices result in almost the same SSIM($I,Y$) after convergence (i.e., when SSIM($I,Y$) doesn't change much).

It is worth noting that the result of basic LHE, $\mathbf{U}$, is not a good choice for $Y_{init}$ as it always results in SSIM($I,Y$) smaller than that resulted by either of the two choices above.

**5. ENHANCING METHOD OF [3] USING DOF**
Since the method of [3] requires calculation of local histograms, it inherits the rank ambiguity, described by (3), that caused the DoFs of LHE solution. In this section, we develop the DoFs of the solution of the main method of [3] (i.e., the one set out in (3), (5), (13), and (16) of [3], when (17) of [3] holds).

The mapping function for signed power-law (SPL) accumulation, using the notation of [3], can be written as

$$z(m,n,g) = \hat{h}(m,n,g) \overset{g}{*} 2^{\alpha-1}\text{sign}(g)\,|g|^\alpha, \tag{16}$$

in which $0 \le \alpha \le 1$ is the only parameter of the method, and $\hat{h}(m,n,.)$ is the local histogram at $(m, n)$. Considering that in [3] the range of input and output GLs is [-0.5, 0.5], the DoF of [3] with SPL accumulation at pixel $(m, n)$ is given by:

$$[z(m,n,I(m,n)-L^{-1})+L^{-1},\, z(m,n,I(m,n))] \tag{17}$$

The main method of [3] requires adding a ratio $\alpha$ (assuming condition (17) of [3] holds) of local mean to the result of SPL mapping. Hence the DoF of the main method of [3] at pixel $(m, n)$ is given by:

$$[z(m,n,I(m,n)-L^{-1})+L^{-1}+\alpha\bar{I}(m,n),\, z(m,n,I(m,n))+\alpha\bar{I}(m,n)], \text{ with} \tag{18}$$

$$\bar{I}(m,n) = I(m,n) \overset{m,n}{*} w^{-2}W(m,n)$$

in which $w$ and $W$ are defined in (1).

To enhance the method of [3], we use the methods of Section 3 (or 4) with the DoFs computed by (18), to achieve MSE- (or SSIM-) optimized results.

## 6. NUMERICAL RESULTS

To compare our methods against the prior art in terms of over-enhancement reduction in the result, we used PSNR or SSIM to measure the similarity between the results and the input image, and the total local energy of the result to measure the contrast. The ideal method should increase both measures.

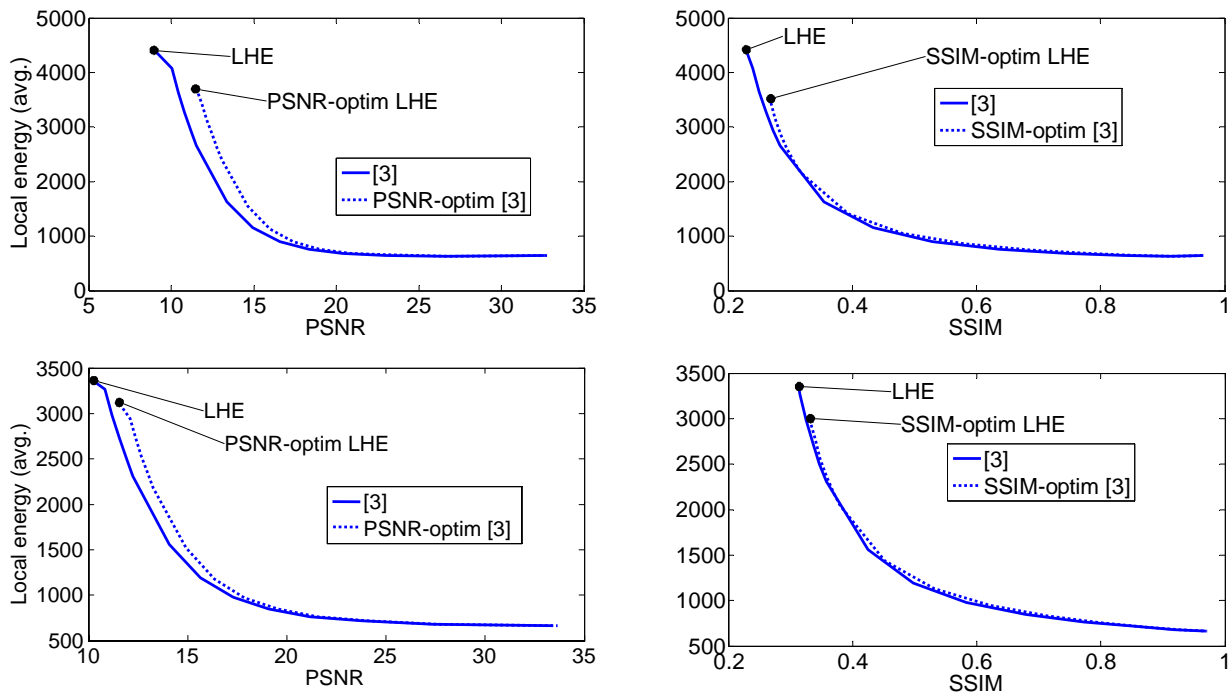

**Fig 1. Top row:** Contrast-similarity trade-off in the results of [3] and its PSNR-optimized (left) and SSIM-optimized (right) versions, for various values of $\alpha$, the parameter of [3]. $\alpha = 0$ corresponds to basic LHE in [3] or optimized LHE in enhanced [3]. A 5x5 neighborhood window is used. **Bottom row:** the same for 17x17 neighborhood.

Fig. 1 shows the trade-off between the measures of similarity and contrast for test image *cameraman* for 5x5 and 17x17 neighborhood windows (i.e., $w = 5$ in (1)) in the result of the method of [3] and its MSE- and SSIM- optimized versions (described in Section 5), for various values of $\alpha$. The points on the graphs show the result of the basic LHE and those of its MSE- and SSIM-optimized versions (described in Sections 3 and 4), which are equivalent to the result of the method of [3] or its enhanced versions for $\alpha = 0$. It is observed that the curves for the enhanced versions are closer to the ideal operating condition (top-right corner of the graphs = high similarity & high contrast). In the case of 17x17 neighborhood window, it is observed that the difference in the performance metrics of [3] and its enhanced versions is smaller than that for 5x5 neighborhood window. That is because as the window size increases, LHE approaches GHE. Hence our method that is based on local optimization becomes less effective.

## 7. CONCLUSIONS

In this paper, we pointed out that the local (adaptive) histogram equalization (or specification) problem has many solutions among which one can be chosen to optimize a certain metric. We gave methods to find LHE solutions more similar to the input image, in MSE or SSIM senses. Such solutions reduce over-enhancement as compared to the existing method.

**APPNEDIX: SSIM AND ITS GRADIENT**

The SSIM index [5] for grayscale images $X$ and $Y$ is defined by

$$\text{SSIM}(X,Y) = \frac{1}{N} \sum_{\forall m,n} \text{SSIM\_map}_{X,Y}(m,n), \tag{A1}$$

where $N$ is the number of pixels in $X$ (and $Y$; the inputs are assumed to be of the same size) and $(m, n)$ are pixel coordinates. $\text{SSIM\_map}_{X,Y}$ (with the same size as $X$) is a map of similarity between $X$ and $Y$. It is larger where $X$ and $Y$ are locally more similar. $\text{SSIM\_map}_{X,Y}$ is given by

$$\text{SSIM\_map}_{X,Y}(m,n) = \frac{(2\mu_X \mu_Y + C_1)(2\sigma_{XY} + C_2)}{(\mu_X^2 + \mu_Y^2 + C_1)(\sigma_X^2 + \sigma_Y^2 + C_2)}, \tag{A2}$$

where $\mu_X$ and $\sigma_X$ denote the local average and local standard deviation of pixels of $X$ (i.e., those within a sliding window centered on $(m, n)$), and $\sigma_{XY} = \overline{(X(i,j) - \mu_X)(Y(i,j) - \mu_Y)}$, in which the average (indicated by bar) is taken over all pixels $(i, j)$ within the sliding windows centered at $(m, n)$ on $X$ and $Y$. Except $C_1$ and $C_2$, the small positive constants that keep the denominator non-zero, all variables on the right side of (A2) are functions of location $(m, n)$ (e.g., $\mu_X$ is actually $\mu_X(m,n)$).

If the input images are identical, the index given by (A1) becomes 1. If they are uncorrelated, the index becomes very small. If one of the inputs is considered the reference image, the index gives the quality of the other input image as compared to the reference.

To be able to compute its gradient, we need to express SSIM in linear terms. To that end, instead of sliding windows (that define local neighborhoods) over the input images to compute the elements of $\text{SSIM\_map}_{X,Y}$ one by one, the $\text{SSIM\_map}_{X,Y}$ can be computed as a whole by substituting

$$\mu_Y = W * Y, \ \sigma_Y^2 = (W * Y^2) - \mu_Y^2, \ \sigma_{XY} = (W * XY) - \mu_X \mu_Y, \ \mu_X = W * X, \ \sigma_X^2 = (W * X^2) - \mu_X^2 \tag{A3}$$

in (A2) and performing element-wise additions, multiplications and divisions. In (A3), $W$ is a 2-D symmetric low-pass kernel and '$*$' denotes 2-D convolution defined by

$$W(m,n) * X(m,n) = \sum_i \sum_j W(i,j) X(m-i, n-j). \tag{A4}$$

Note that all variables in (A3) are functions of location (i.e., 2-D signals; $(m, n)$'s are removed) and that except the convolution, all operations are performed element-wise. To be identical to the sliding window computation of SSIM, with a $(2w+1) \times (2w+1)$ window, the kernel $W$ in (A3) must be

$$W(m,n) = \begin{cases} (2w+1)^{-2}, & |m| \leq w \text{ and } |n| \leq w \\ 0, & \text{Otherwise} \end{cases} \tag{A5}$$

(i.e., 2-D moving average filter). In [5], a Gaussian kernel with standard deviation of 1.5 (truncated to 11x11 & normalized) is suggested for to the best conformity to perceptual quality.

The SSIM gradient with respect to its input image $Y$ is given by [6]:

$$\nabla_Y \text{SSIM}(X,Y) = \frac{1}{N} \left( W * F + \left( W * \frac{\partial \text{SSIM}_{\text{map}}}{\partial \sigma_{XY}} \right) X + 2 \left( W * \frac{\partial \text{SSIM}_{\text{map}}}{\partial \sigma_Y^2} \right) Y \right), \text{ with} \tag{A6}$$

$$F = \frac{2\mu_X (2\sigma_{XY} + C_2 - 2\mu_X \mu_Y - C_1) - 2\mu_Y (\sigma_X^2 + \sigma_Y^2 + C_2 - \mu_X^2 - \mu_Y^2 - C_1) \text{SSIM}_{\text{map}}}{D(X,Y)},$$

$$\frac{\partial \text{SSIM}_{\text{map}}}{\partial \sigma_{XY}} = \frac{2(C_1 + 2\mu_X \mu_Y)}{D(X,Y)}, \text{ and } \frac{\partial \text{SSIM}_{\text{map}}}{\partial \sigma_Y^2} = \frac{-\text{SSIM}_{\text{map}}}{\sigma_X^2 + \sigma_Y^2 + C_2},$$

in which $\text{SSIM}_{\text{map}}$ is given by (A2), and $D(X,Y) = (\mu_X^2 + \mu_Y^2 + C_1)(\sigma_X^2 + \sigma_Y^2 + C_2)$ is its denominator.